\newif\ifdebug
\documentclass[sigconf]{acmart}

\usepackage{svg}
\usepackage{graphicx}
\usepackage{xspace}
\usepackage[nameinlink]{cleveref}
\usepackage{multirow}
\usepackage{todonotes}
\usepackage{layouts}
\usepackage{csquotes}
\usepackage{marginnote}
\usepackage{enumitem}
\AtBeginDocument{%
  }

\copyrightyear{2026}
\acmYear{2026}
\setcopyright{cc}
\setcctype{by}
\acmConference[SIGIR '26]{Proceedings of the 49th International ACM SIGIR Conference on Research and Development in Information Retrieval}{July 20--24, 2026}{Melbourne, VIC, Australia}
\acmBooktitle{Proceedings of the 49th International ACM SIGIR Conference on Research and Development in Information Retrieval (SIGIR '26), July 20--24, 2026, Melbourne, VIC, Australia}
\acmDOI{10.1145/3805712.3809576}
\acmISBN{979-8-4007-2599-9/2026/07}

\title{Is It Novel and Why? Fine-Grained Patent Novelty Prediction Based on Passage Retrieval}

\author{Valentin Knappich}
\affiliation{%
  \institution{Bosch Center for AI}
  \city{Stuttgart}
  \country{Germany}
}
\affiliation{%
  \institution{University of Augsburg}
  \city{Augsburg}
  \country{Germany}
}
\email{valentin.knappich@de.bosch.com}

\author{Anna H\"atty}
\affiliation{%
  \institution{Bosch Center for AI}
  \city{Stuttgart}
  \country{Germany}
}
\email{anna.haetty@de.bosch.com}

\author{Simon Razniewski}
\affiliation{%
  \institution{ScaDS.AI \& TU Dresden}
  \city{Dresden}
  \country{Germany}
}
\email{simon.razniewski@tu-dresden.de}

\author{Annemarie Friedrich}
\affiliation{%
  \institution{University of Augsburg}
  \city{Augsburg}
  \country{Germany}
}
\email{annemarie.friedrich@uni-a.de}

\renewcommand{\shortauthors}{Knappich et al.}

\ccsdesc[500]{Computing methodologies~Information extraction}
\ccsdesc[500]{Computing methodologies~Natural language generation}
\ccsdesc[500]{Computing methodologies~Search methodologies}

\keywords{Large Language Models, LLM, Passage Retrieval, Patent Analysis, Novelty Prediction, Explainable AI}

\definecolor{anne}{rgb}{0,0.5,0.9}
\definecolor{valentin}{rgb}{0.998,0.722,0.635}
\definecolor{anna}{rgb}{0,0.5,0.9}
\definecolor{simon}{rgb}{0.998,0.722,0.635}

\definecolor{darkgreen}{rgb}{0.0, 0.5, 0.0}

\ifdebug
    \paperwidth=\dimexpr \paperwidth + 20cm\relax
    \oddsidemargin=\dimexpr\oddsidemargin + 10cm\relax
    \evensidemargin=\dimexpr\evensidemargin + 10cm\relax
    \marginparwidth=\dimexpr \marginparwidth + 10cm - \marginparsep\relax

    \newcommand{\valentintodo}[2][valentin]{\todo[color=#1,size=\footnotesize]{\textbf{VK:} #2}}
    \newcommand{\annetodo}[2][anne!30]{\todo[color=#1,size=\footnotesize]{\textbf{AF:} #2}}
    \newcommand{\annatodo}[2][anna]{\todo[color=#1,size=\footnotesize]{\textbf{AH:} #2}}
    \newcommand{\simontodo}[2][simon!30]{\todo[color=#1,size=\footnotesize]{\textbf{SR:} #2}}

    \newcommand{\valentintodofigure}[2][0cm]{\marginnote{\todo[color=valentin,size=\footnotesize,inline]{\textbf{VK:} #2}}[#1] }
    \newcommand{\annetodofigure}[2][0cm]{\marginnote{\todo[color=anne!30,size=\footnotesize,inline]{\textbf{AF:} #2}}[#1] }
    \newcommand{\annatodofigure}[2][0cm]{\marginnote{\todo[color=anna,size=\footnotesize,inline]{\textbf{AH:} #2}}[#1] }
    \newcommand{\simontodofigure}[2][0cm]{\marginnote{\todo[color=simon!30,size=\footnotesize,inline]{\textbf{SR:} #2}}[#1] }

\else
    \newcommand{\valentintodo}[2][valentin]{}
    \newcommand{\annetodo}[2][anne!30]{}
    \newcommand{\annatodo}[2][anna]{}
    \newcommand{\simontodo}[2][simon!30]{}

    \newcommand{\valentintodofigure}[2][0cm]{}
    \newcommand{\annetodofigure}[2][anne!30]{}
    \newcommand{\annatodofigure}[2][anna]{}
    \newcommand{\simontodofigure}[2][simon!30]{}

\fi

\begin{document}

\newcommand{\ul}[1]{\underline{#1}}
\def\DatasetNameLong{%
  \ul{Fi}ne-grained
  \ul{N}ovelty
  \ul{E}xamination
  of Patents%
}
\def\DatasetName{\textsc{FiNE-Patents}\xspace}

\def\mathdefault#1{#1}

%%
%% The abstract is a short summary of the work to be presented in the
%% article.
\begin{abstract}
Novelty assessment is a critical yet complex task in the examination process for patent acceptance, requiring examiners to determine whether an invention is disclosed in a prior art document.
The process involves intricate matching between specific features of a patent claim and passages in the prior art. 
While prior work has approached novelty prediction primarily as a binary classification task at the claim level, we argue that this formulation is susceptible to spurious correlations and lacks the granularity required for practical application.
In this work, we introduce \DatasetName (\DatasetNameLong), a novel dataset comprising 3,658 first patent claims annotated with fine-grained, feature-level prior art references extracted from European Search Opinion (ESOP) documents.
We propose shifting the evaluation paradigm from simple binary classification to a joint retrieval and abstract reasoning task at the feature level, requiring models to identify specific passages from a prior art document that disclose individual claim features, and to identify which features of a claim make it novel.
We implement and evaluate LLM-based workflows that decompose claims into features, analyze each feature against prior art, and finally derive a claim-level novelty prediction.
Our experiments demonstrate that these workflows outperform embedding-based baselines on passage retrieval and novel feature identification. 
Furthermore, we show that unlike trained classifiers, LLMs are robust against spurious correlations present in the claim-level novelty classification task. 
We release the dataset and code to foster further research into transparent and granular patent analysis.
\end{abstract}

%% A "teaser" image appears between the author and affiliation
%% information and the body of the document, and typically spans the
%% page.
% \begin{teaserfigure}

%   \Description{TODO}
%   \label{fig:teaser}
% \end{teaserfigure}

%%
%% This command processes the author and affiliation and title
%% information and builds the first part of the formatted document.
\maketitle

\begin{figure}[t!]
  \centering
  \includegraphics[width=\linewidth]{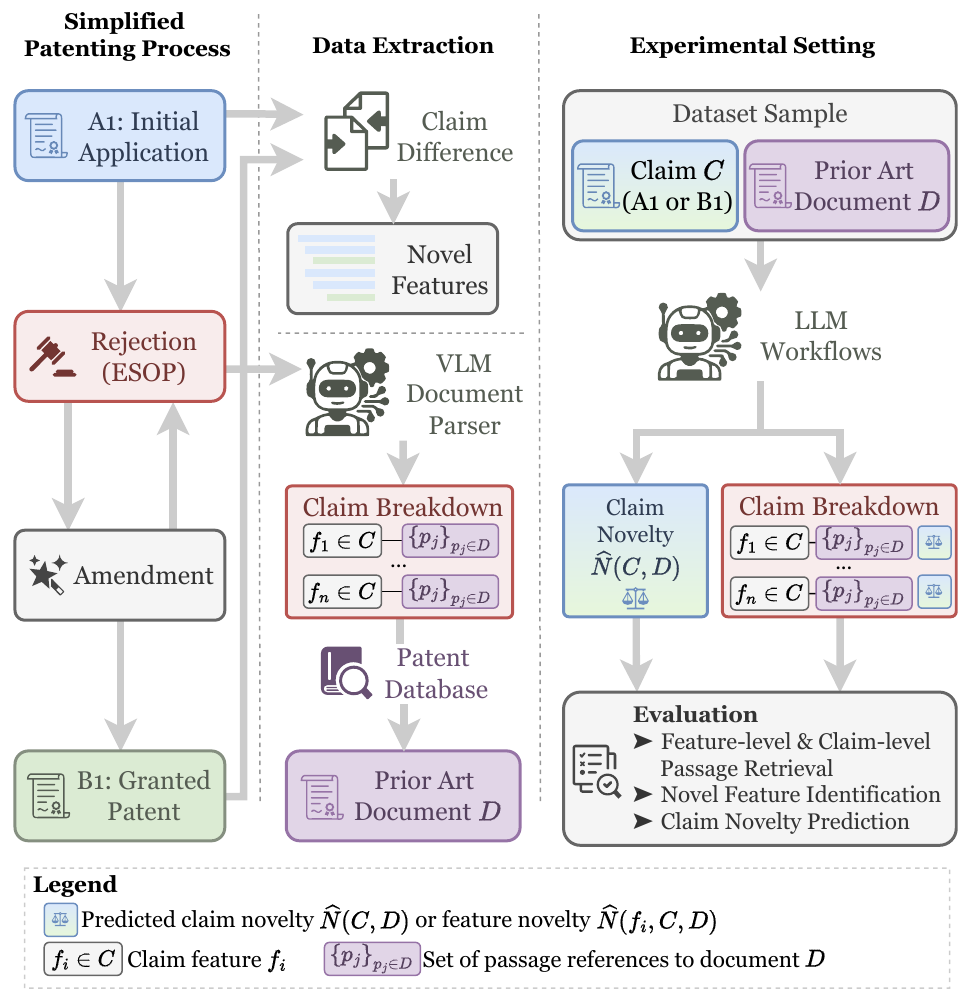}
  \caption{
    Methodology Overview. [Left] Simplified Patenting Process: The initial application (A1) is rejected due to lack of novelty in a European Search Opinion (ESOP) document and amended (typically at least once) until the granted patent (B1) is issued. [Middle] Data Extraction: We extract feature-level prior art references from the ESOP and retrieve the prior art document cited as the closest prior art $D$. [Right] Experimental Setting: Given a claim $C$ (either from the initial or the granted version) and the closest prior art document, the model retrieves relevant paragraphs $\{ p_j \}_{p_j \in D}$ per claim feature $f_i$, determines whether each feature is disclosed in $D$, and predicts whether the claim $C$ is novel over $D$.
  }
  \Description{
    Methodology Overview. [Left] Simplified Patenting Process: The initial application (A1) is rejected due to lack of novelty in a European Search Opinion (ESOP) document and amended (typically at least once) until the granted patent (B1) is issued. [Middle] Data Extraction: We extract feature-level prior art references from the ESOP and retrieve the prior art document cited as the closest prior art $D$. [Right] Experimental Setting: Given a claim $C$ (either from the initial or the granted version) and the closest prior art document, the model retrieves relevant paragraphs $\{ p_j \}_{p_j \in D}$ per claim feature $f_i$, determines whether each feature is disclosed in $D$, and predicts whether the claim $C$ is novel over $D$.
  }
  \label{fig:teaser}
\end{figure}

\section{Introduction}\label{sec:intro}

\begin{figure*}
  \centering
  \includegraphics[width=\textwidth]{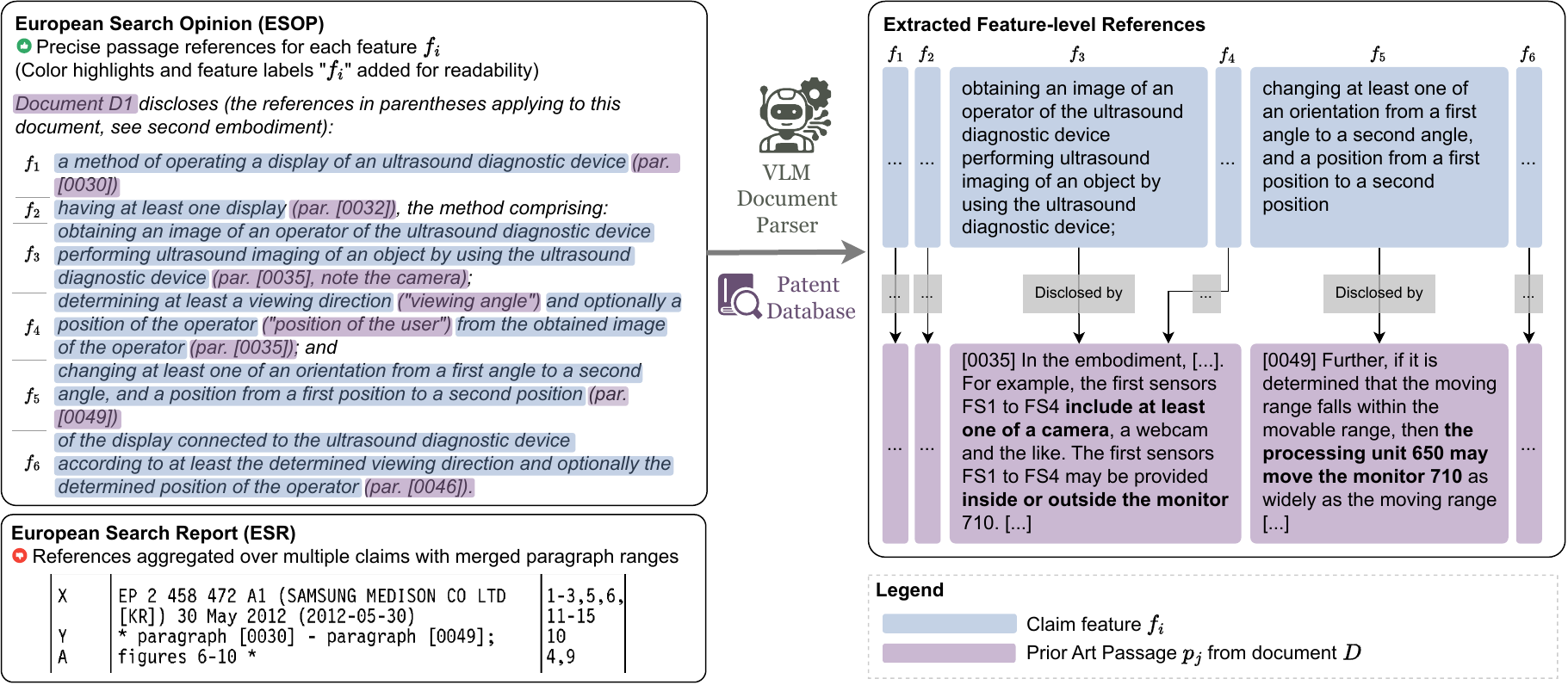}
  \caption{
    Example of the feature-level prior art references. [Top left] In the ESOP document, the examiner rejects claim 1 for lack of novelty over prior art document $D$ (here: D1) and provides feature-level references to $D$ (claim features highlighted in blue, references in purple). [Bottom left] The European Search Report (ESR) commonly used by prior work lists references aggregated over multiple claims with merged paragraph ranges and without mapping passages to features. [Right] The extracted feature-level references for claim 1, showing which passages in $D$ disclose which features of the claim. 
  }
  \Description{
    Example of the feature-level prior art references. [Top left] In the ESOP document, the examiner rejects claim 1 for lack of novelty over prior art document $D$ (here: D1) and provides feature-level references to $D$ (claim features highlighted in blue, references in purple). [Bottom left] The European Search Report (ESR) commonly used by prior work lists references aggregated over multiple claims with merged paragraph ranges and without mapping passages to features. [Right] The extracted feature-level references for claim 1, showing which passages in $D$ disclose which features of the claim. 
  }
  \label{fig:example}
\end{figure*}

Patents play a crucial role in protecting inventions and fostering innovation across various industries.
Computational support for the patent domain has been actively studied for decades, with particularly strong activity in prior-art search and patent classification \citep{lupu2017patent,piroi2017evaluating,krestel2021survey,asi7050091,shomee-etal-2025-survey,CHI2026114063}.
Patent novelty prediction has recently received increased attention \cite{jiang2025natural,ikoma2025aiexaminenoveltypatents,lim2025panorama,10.1145/3726302.3729970} and aims to predict whether the invention defined by a patent has already been disclosed in prior work.
Within patents, the set of claims defines the scope of the legal protection sought by the applicant.
Each claim is a single sentence defining the invention as a combination of multiple \textit{features}, where each feature is a component, step, or characteristic of the invention.
For example, in a claim describing a new text classification method, individual features could include $f_1 = $ \textit{converting text into tokens}, $f_2 = $ \textit{applying a neural network to the tokens}, and $f_3 = $ \textit{mapping embeddings into the label space}.
An invention is considered novel over a \textit{prior art document} if at least one feature of the claim is not disclosed in the prior art document.
Predicting patent novelty thus requires comparing each feature of a claim against the prior art, demanding both deep technical understanding and highly abstract reasoning.

Prior work has primarily evaluated novelty prediction at the claim level as a binary classification task \cite{risch2021patentmatch,Blume2024ComparingCC,Parikh2024ClaimCompareAD,ikoma2025aiexaminenoveltypatents,lim2025panorama}.
However, in most setups, this is highly susceptible to spurious correlations leading to overly optimistic results of trained models \cite{Blume2024ComparingCC,ikoma2025aiexaminenoveltypatents}, even after stratifying for claim length and removing the strongest biases.
To analyze the spurious correlations, we create an adversarial test set using adversarial filtering \cite{le2020adversarial} to detect whether models make use of them.
Interestingly, we find that zero-shot Large Language Models (LLMs) do not rely on these shortcuts, achieving similar performance on the adversarial test set as on the regular test set.

Crucially, we argue that patent novelty examination should also be evaluated at the feature level.
A claim novelty label provides little to no insight into the specific aspects of the invention that might be disclosed by a prior art document, making it of limited practical use.
Instead, we propose a more nuanced evaluation protocol that requires models to (i) identify which passages are relevant for a given feature (\textit{passage retrieval}), (ii) determine which features are novel (\textit{novel feature identification}), and (iii) predict whether the claim is novel (\textit{claim novelty prediction}).
This approach provides patent professionals with more actionable and trustworthy support, as they can transparently trace and verify the model's reasoning by directly inspecting the cited source passages underlying each identified match.
\Cref{fig:teaser} provides an overview of our methodology.

To support this evaluation framework, we introduce the \DatasetName dataset based on examination records from the European Patent Office (EPO).
In the European patent system, the initial application document (referred to as \textit{A1}) undergoes examination and is typically amended one or more times in response to examiner objections before a patent is finally granted (referred to as \textit{B1}).\footnote{\url{https://register.epo.org/help?lng=en&topic=kindcodes}}
When a claim lacks novelty, the examiner writes a European Search Opinion (ESOP) document explaining the rejection.
The ESOP cites the most relevant prior art document (referred to as $D$) and provides a detailed \textit{breakdown} of the claim into features, describing which passages of $D$ disclose which features of the claim.
We extract these feature-level prior art references from the ESOP documents to evaluate models' novelty analyses.
An example of the extraction process is shown in \Cref{fig:example}.

To generate detailed novelty analyses, we implement workflows using LLMs to output both claim-level novelty predictions and feature-level passage references.
We use either just a single LLM call, or analyze each feature separately and then aggregate the results.
We find that our methods outperform baselines based on embedding similarity on retrieval metrics and that they do not rely on shortcuts based on spurious correlations for novelty prediction.

We summarize our contributions as follows:
\begin{enumerate}
  \item We propose to evaluate patent novelty examination across three sub-tasks: \textit{passage retrieval} at the feature and claim level, \textit{novel feature identification}, and claim-level \textit{novelty prediction}.
  
  \item We present \DatasetName, a dataset of 3,658 first patent claims with feature-level prior art passage references and novelty labels.

  \item We implement LLM-based workflows to predict patent novelty that split the claims into features, analyze each feature separately, and then aggregate the results.
  
  \item We publicly release the dataset, code to create the dataset, and code to reproduce our experiments.\footnote{\url{https://github.com/boschresearch/fine-patents}}
\end{enumerate}

\section{Related Work}\label{sec:related_work}

Novelty assessment is a fundamental challenge across multiple domains, from evaluating creative ideas \cite{beaty2021automating,organisciak2023beyond,saretzki2026investigating} and scientific research contributions \cite{lin-etal-2025-evaluating,ai-etal-2025-novascore,zhao2025review,Wu2025SC4ANMIO,10.1145/3527546.3527577,10.1145/2911451.2911498} to assessing patent applications.
In the patent domain, recent advances in large language models have opened new possibilities for supporting various tasks, including prior art search \cite{ding2025automatic,10.1145/3726302.3729970}, patent drafting \cite{wang2024autopatent,knappich-etal-2025-pap2pat}, analysis \cite{shomee-etal-2025-survey,krestel2021survey,wang2024evopat,knappich2025pedantic}, and novelty prediction \cite{hido2012modeling,lo2021pre,shan2024structural,risch2021patentmatch,Blume2024ComparingCC,Parikh2024ClaimCompareAD,Shen2024PatentGrapherAP,lee2024patenteditsframingpatentnovelty,ikoma2025aiexaminenoveltypatents,lim2025panorama,10.1145/3726302.3729970}.
Our work builds upon this emerging body of research by introducing fine-grained, feature-level supervision for patent novelty examination.

\paragraph{Patent Grant Prediction}

Predicting the likelihood of a patent application being granted has been one of the most prevalent tasks in patent NLP \cite{jiang2025natural}.
Prior works have either predicted the final grant decision as a whole \cite{hido2012modeling,lo2021pre, shan2024structural}, or separately predicted novelty and/or inventiveness \cite{risch2021patentmatch,Blume2024ComparingCC,Parikh2024ClaimCompareAD,Shen2024PatentGrapherAP,lee2024patenteditsframingpatentnovelty,ikoma2025aiexaminenoveltypatents,lim2025panorama,10.1145/3726302.3729970}, lack of definiteness \cite{knappich2025pedantic}, or insufficient enablement \cite{kong2023linguistic,yoo2026pat}. \citet{lim2025panorama} introduce the PANORAMA dataset for patent novelty and inventiveness prediction, which contains claim-level passage references from US rejection documents.

\paragraph{Patent Passage Retrieval}

The retrieval of relevant patent prior art passages has been studied in various works \cite{fujii2005document,andersson2013exploring,andersson2016time,ayaou2025dapfam}, most notably based on the CLEF-IP 2012 \cite{Andersson2012ReportOT}, CLEF-IP 2013 \cite{piroi2013overview}, and NTCIR 2005 \cite{Fujii2005OverviewOP,fujii2006test} datasets.
They study the retrieval of relevant passages from a large corpus of patent documents given a query claim.
In contrast, our work focuses on finding relevant passages from a single prior art document and uses feature-level labels extracted from office action full texts.

\section{\DatasetName Dataset}\label{sec:dataset}

\begin{table}[!t]

\setlength{\tabcolsep}{4pt}

\begin{tabular}{lll}
    \toprule
    Statistic & & Count\\
    \midrule
    \multirow{5}{*}{\# Claims} & All & 3658;\phantom{aaaaaaa} 50.0\% novel \\
    & Train & 1454 (39.7\%); 50.1\% novel \\
    & Val & 360 (9.8\%);\phantom{aa} 52.2\% novel \\
    & Test & 1844 (50.4\%); 49.5\% novel \\
    & \textcolor{gray}{Adversarial Test} & \textcolor{gray}{278 (7.6\%);\phantom{aa} 50.0\% novel} \\
    \midrule
    \multirow{4}{*}{\# Applications} & All & 3163 \\
    & Rejected claim only & 1334 (42.2\%) \\
    & Granted claim only & 1334 (42.2\%) \\
    & Both claims & 495 (15.6\%) \\
    \midrule
    \multirow{2}{*}{\shortstack{\# Features\\per Claim}} & All & 6.2 $\pm$ 2.4 \\
    & w/ References & 4.5 $\pm$ 2.2 \\
    \midrule
    \multirow{6}{*}{Cited Patents} & All & 3658 \\
    & US & 2658 (72.7\%) \\
    & EP & 996 (27.2\%) \\
    \cmidrule{2-3}
    & \# paragraphs & 115.9 $\pm$ 90.7 \\
    & \# claims & 21.9 $\pm$ 23.4 \\
    & \# words & 11662.8 $\pm$ 8963.4 \\
    \midrule
    \multirow{6}{*}{\shortstack{\# Cited\\Passages}} & Total & 21464 \\
    & Per Claim & 11.7 $\pm$ 25.0 \\
    & Per Feature & 1.9 $\pm$ 4.4 \\
    \cmidrule{2-3}
    & Paragraph & 20166 (94.0\%) \\
    & Claim & 1020 (4.8\%) \\
    & Abstract & 278 (1.3\%) \\
    \bottomrule
  \end{tabular}

  \caption{Dataset Statistics. Averages are reported as mean $\pm$ standard deviation. \textit{Adversarial Test} is a subset of \textit{Test}.}
  \label{tab:dataset_stats}
\end{table}

In this section, we describe the \DatasetName dataset, whose statistics are summarized in \Cref{tab:dataset_stats}.
The dataset contains examination data from 3,163 patent applications filed at the EPO between 2012 and November 2025, comprising 3,658 first claims with an even split between novel and not novel claims.
For each claim, the dataset includes the first ESOP document, written by the examiner and parsed into a structured format, as well as the full text of the closest prior art document cited in the ESOP.
In the following, we provide details on the task definition and evaluation protocol (\Cref{sec:eval_protocol}), the data source (\Cref{subsec:search_report_vs_search_opinion}), the dataset creation process (\Cref{sec:dataset_creation}), an analysis of spurious correlations (\Cref{sec:dataset_spurious}), and an analysis of data contamination (\Cref{sec:data_contamination}).

\subsection{Task Definitions and Evaluation Protocol}\label{sec:eval_protocol}

We propose to evaluate models on three sub-tasks: retrieval of relevant prior art passages from the closest prior art document, the identification of novel features in a claim, and claim novelty prediction.
The input is a claim under examination $C$ and its closest prior art document $D$ consisting of passages $p_j \in D$.
Each claim $C$ consists of features $f_i$ as defined by a segmentation $S(C) = \{f_1, f_2, \ldots, f_n\}$.
A gold feature corresponds to the span of the claim for which the examiner has provided references in the ESOP document.
For generality, we do not assume access to the gold segmentation at inference time, i.e., models must first produce their own segmentation $\hat{S}(C)$.

\subsubsection{Passage Retrieval}

The first sub-task is the retrieval of relevant passages from the prior art document $D$ that disclose the features $f_i$ of the claim $C$.
That is, for each feature $f_i \in \hat{S}(C)$, the full claim $C$, and the prior art document $D$, the model must retrieve passages $p_j$ that disclose $f_i$.
We refer to the set of passages cited in the ESOP as $P(f_i, C, D) = \{p_j \in D \mid p_j \text{ discloses } f_i\}$ and to the set of passages retrieved by the model as $\hat{P}(f_i, C, D)$.
For evaluation at the feature level, we compute the precision (P), recall (R), F1, and nDCG \cite{DBLP:journals/tois/JarvelinK02} scores between $P(f_i, C, D)$ and $\hat{P}(f_i, C, D)$ by mapping $f_i \in \hat{S}(C)$ to the feature in $S(C)$ with minimal edit distance.
At the claim level, we aggregate the passages over all features $P(C, D) = \bigcup_{f_i \in S(C)} P(f_i, C, D)$, i.e., $P(C, D)$ contains all passages that disclose at least one feature of the claim.

While cited passages are reliably indicative of disclosure, the absence of a citation does not necessarily imply non-disclosure, i.e., negatives may not be true negatives.
To mitigate this issue, we additionally report soft variants of precision, recall, and F1 (\~P, \~R, \~F1) \cite{franti2023soft}.
To obtain them, we compute the ROUGE-L \cite{lin-2004-rouge} score between each predicted passage and each cited passage, and aggregate:

\begin{align*}
\text{\~P} = \frac{1}{|\hat{P}(f_i, C, D)|} \sum_{p_j \in \hat{P}(f_i, C, D)} \max_{p_k \in P(f_i, C, D)} \text{ROUGE-L}(p_j, p_k)\\
\text{\~R} = \frac{1}{|P(f_i, C, D)|} \sum_{p_k \in P(f_i, C, D)} \max_{p_j \in \hat{P}(f_i, C, D)} \text{ROUGE-L}(p_j, p_k)
\end{align*}

This can be seen as a soft generalization of precision and recall, where each predicted passage receives partial credit based on its maximum overlap with any ground truth passage (for precision), and symmetrically, each ground truth passage is scored by its maximum overlap with any predicted passage (for recall).
We argue that lexical overlap is a reasonable proxy for soft labels: while matching claim features to prior art passages across different applications requires bridging the vocabulary gap, the terminology used within a single prior art document is largely consistent, making ROUGE-L a suitable similarity measure.
Since ESOP documents always refer to the initially filed version of the claim, we compute retrieval scores only for the claims labelled as not novel.

\subsubsection{Novel Feature Identification}

At the core of novelty assessment is the identification of novel features.
Given the model's binary novelty prediction $\hat{N}(f_i, C, D) \in \{0, 1\}$ for each feature $f_i \in \hat{S}(C)$, we extract the character ranges of features predicted as novel and compare them to the character ranges actually added during prosecution.
We report precision (P), recall (R), and F1 scores.
Since added features are only available for the granted version of the claims, we compute these scores only for the claims labelled as novel.

\subsubsection{Claim Novelty Prediction}

The final sub-task is the binary classification of whether the claim $C$ is novel over the prior art document $D$, referred to as $\hat{N}(C, D)$.
We report the fraction of claims predicted to be novel, accuracy, and macro F1.

\subsection{Data Source}\label{subsec:search_report_vs_search_opinion}

Prior works have utilized the paragraph reference information from \textit{European Search Reports (ESRs)} to construct patent novelty datasets \cite{risch2021patentmatch,Blume2024ComparingCC}.
An example is shown in the bottom left of \Cref{fig:example}.
ESRs list the relevant prior art documents, categorize them by their relevance to novelty and inventive step,\footnote{\url{https://www.epo.org/en/legal/guidelines-pct/2025/b_x_9_2.html}} and reference the relevant passages in the prior art.
They are, unlike ESOPs, available at scale in a structured format. 
However, the references are often very coarse-grained.
By comparing our parsed ESOP citations to the ESRs, we find that only 19\% of the passages cited in the ESR are also cited in the ESOP, indicating a large discrepancy between the two sources of supervision.
Upon inspection of examples, we find that this discrepancy has two main causes.
First, the ESR citations are often aggregated over multiple claims, losing information about which passage relates to which claim.
Second, examiners often merge ranges for brevity, e.g., citing paragraphs \texttt{10-20} instead of \texttt{10-13, 16-18, 20}.
In contrast, ESOPs provide precise references at the feature level, and are thus the superior source of supervision at the cost of expensive PDF parsing.

\subsection{Dataset Creation}\label{sec:dataset_creation}

In the following, we describe the dataset creation process, which consists of four main steps: (i) retrieval of seed patent applications, (ii) parsing of ESOP documents to extract feature-level prior art references, (iii) retrieval of the full text of the cited prior art documents, and (iv) extraction of newly added features from the initial to the granted claim.

\subsubsection{Seed Patent Applications}

We start the dataset creation process by retrieving all EPO patent applications from the CPC classes
\texttt{G06}\footnote{G06: Computing; Calculating; Counting},
\texttt{G10L}\footnote{G10L: Speech Analysis Techniques Or Speech Synthesis; Speech Recognition; Speech Or Voice Processing Techniques; Speech Or Audio Coding Or Decoding}, \texttt{H04L}\footnote{H04L: Transmission Of Digital Information}, and
\texttt{H04N}\footnote{H04N: Pictorial Communication} from 2012 onwards.
These classes were chosen such that the information retrieval community is familiar with the domain of the patents, making it easier to understand the examination.
To exclude low-quality applications, we filter for applications that were finally granted.
Patent application and publication numbers are retrieved from the EPO's Linked Data service\footnote{\url{https://data.epo.org/linked-data/about}} and publication full texts are retrieved from the EPO's Publication Server API\footnote{\url{https://www.epo.org/en/searching-for-patents/technical/publication-server}}.

\subsubsection{ESOP Parsing}

For each seed patent application, we download, if available, the first \textit{European Search Opinion} document from the EPO register.\footnote{\url{https://register.epo.org/}}
These documents contain the examiner's reasoning for rejecting a subset of the claims.
For rejections based on lack of novelty, the examiner recites the rejected claim and provides precise references to the closest prior art document $D$ for every feature of the claim.
Our analysis centers on the first claim of each application, as it is generally the broadest and most significant independent claim, and is most frequently examined in depth within the ESOP.
We extract these feature-level prior art references using the vision language model (VLM) \texttt{Qwen3\--VL\--235B\--A22B\--Thinking} \cite{Qwen3-VL} with entirely image-based input and structured output.
This end-to-end approach avoids the accumulation of errors in multi-step pipelines including OCR and yields highly accurate results, but it is costly to scale.
To minimize costs, we first filter for documents that contain a breakdown of the first claim using the highly efficient OCR model MinerU 2.5\footnote{\url{https://huggingface.co/opendatalab/MinerU2.5-2509-1.2B}} \cite{niu2025mineru2} and regular expressions that match the claim text with additional references in parentheses.
\Cref{fig:example} shows an example of an ESOP document on the left and a visualization of the structured output on the right.

\begin{figure}[!t]
  \centering
  \input{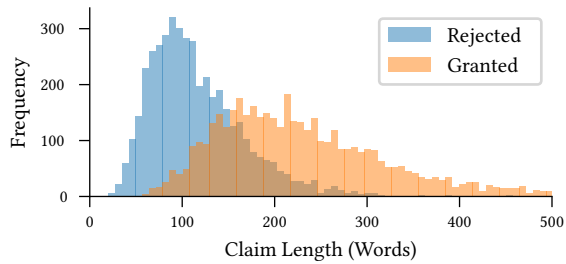}
  \caption{Histogram of claim lengths before stratification. The length of the claim under examination provides a strong shortcut for predicting novelty. After stratified sampling, each class has the distribution of the overlap.}
  \Description{Histogram of claim lengths before stratification. The length of the claim under examination provides a strong shortcut for predicting novelty. After stratified sampling, each class has the distribution of the overlap.}
  \label{fig:claim_lens}
\end{figure}

\subsubsection{Prior Art Documents}

For each prior art document cited in the ESOP, we retrieve the full text from the EPO Publication Server API for all available jurisdictions. 
For U.S. patents cited as prior art, we obtain the full text from the USPTO bulk data dump.\footnote{\url{https://bulkdata.uspto.gov}}
To limit the complexity of the pipeline, we only consider prior art patents filed at the EPO or USPTO, leaving the inclusion of patents from other jurisdictions and non-patent literature to future work.

\subsubsection{Newly Added Features}

To evaluate novel feature identification, we leverage the edit operations performed by applicants during prosecution.
Since applicants typically add new limitations to overcome novelty objections, character spans added between the initial and granted claim serve as a strong proxy for novel features.
We determine the newly added character spans in the finally granted claim in B1 compared to the initially filed claim in A1 using a simple diff algorithm based on character-level Levenshtein distance \cite{levenshtein1966binary}.
That is, we determine the character ranges in the claim that were added during prosecution.
To evaluate, we compute the overlap of these ranges with the features identified as novel by a model as described above.

\begin{figure*}[!t]
  \centering
  \includegraphics[width=\linewidth]{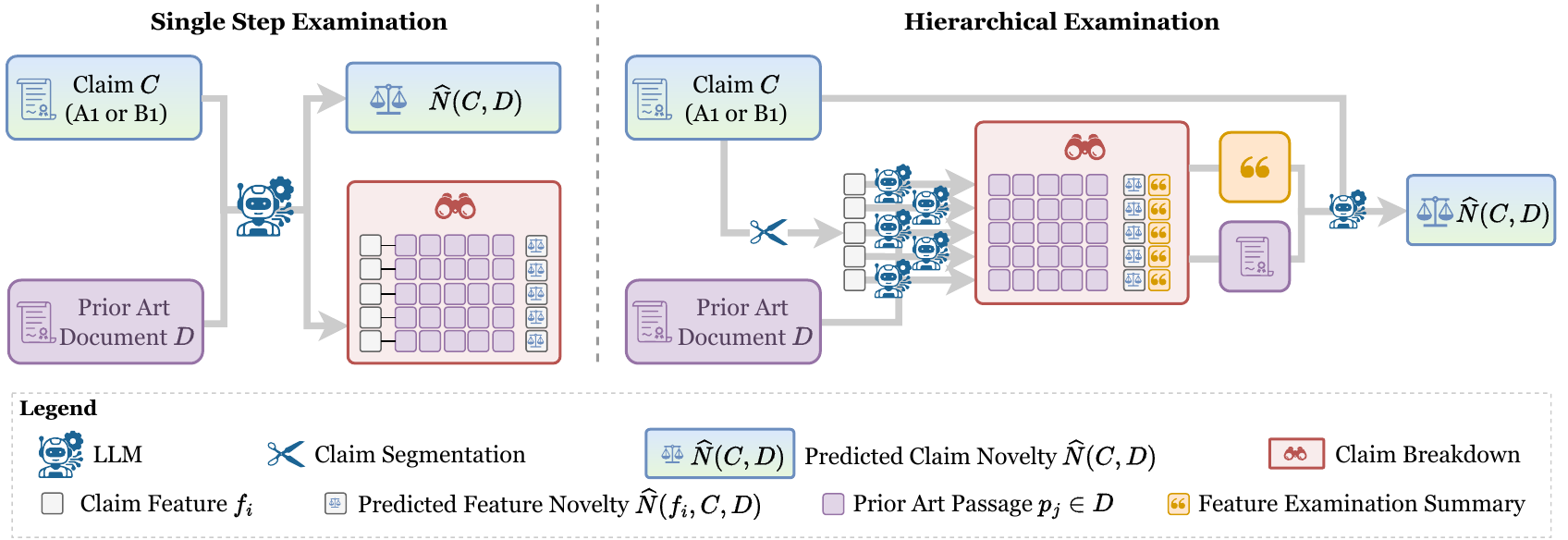}
  \caption{
    LLM Workflows. [Left] In \textit{Single Step Examination}, the model receives the claim and full prior art document, and outputs the breakdown and claim novelty prediction. [Right] In \textit{Hierarchical Examination}, the claim is first segmented into features, each feature is examined separately against the prior art document, and finally, the results are aggregated into a claim novelty prediction.
  }
  \Description{Illustration of the two LLM workflows: Single Step Examination and Hierarchical Examination}
  \label{fig:workflows}
\end{figure*}

\subsubsection{Filtering Statistics}

The dataset creation described above represents a complex pipeline with multiple steps, at each of which a percentage of the samples is discarded.
The initial set of seed applications from the selected CPC classes contained 364k patent applications, out of which 132k were ultimately granted.
Of these granted applications, 67k have complete publication full texts available in the EPO Publication Server.
65k additionally have at least one ESOP document available.
Of these 65k ESOP documents, 19k recite the full first claim with feature-level prior art references, of which 17k are successfully parsed using our VLM document parser.
After filtering out rejections due to lack of inventive step, samples with unavailable prior art documents and samples with unlocatable references, 4.4k applications remain.
Since each application contributes both an initial and a granted claim, this amounts to 8.8k claims.
We additionally perform a claim length stratification as described below, after which 3,658 claims remain.
Lastly, we split the dataset into training, validation, and test sets with a 40/10/50 ratio.

\subsection{Spurious Correlation Mitigation}\label{sec:dataset_spurious}

A common concern is that models may exploit spurious correlations in the dataset to achieve high classification performance without performing the intended task \citep{geirhos2020shortcut}.
In our setup, the claim-level novelty prediction task is particularly susceptible to this issue, because there are superficial differences between the initial and granted claims.
In the following, we describe our efforts to mitigate spurious correlations in the dataset and analyze the remaining ones.

\subsubsection{Length Stratification}

As prior work has noted, the claim length is a strong signal for novelty (\cite{ikoma2025aiexaminenoveltypatents}, see \Cref{fig:claim_lens}) because applicants typically add limitations to overcome novelty objections. 
To mitigate this effect, we stratify the dataset based on the length of the claim into 100 bins and sub-sample the majority class, such that all bins contain an equal number of novel and not novel samples.
\Cref{fig:claim_lens} shows the claim length distribution before stratification.
As a result of our sampling procedure, both classes share the same claim length distribution, shown as the overlapping region in the figure.

\subsubsection{Reference Numerals and Patent Numbers}

Patent documents often use reference numerals to refer to specific parts of the invention across the claims, description, and drawings (e.g., ``a device (10) comprising...'').
Upon inspection of the dataset, we find that they are often omitted in the initial application and added during prosecution. 
On average, granted claims contain 12.1 numerals while initial claims contain only 0.3.
Since they are easy to detect with regular expressions, we remove them from all claims in the dataset.

We additionally note that the descriptions of granted patents contain more patent number references than those of initial applications (2.1 vs.\ 0.2 on average).
In particular, the closest prior art document cited in the ESOP is referenced 0.75 times on average in granted patents but almost never in initial applications.
Since we do not provide the description of the application under examination to the models, this does not affect our experimental setup but should be considered in settings that include the description as input.

\subsubsection{Adversarial Test Set}

While the above steps mitigate the strongest shortcuts, more subtle spurious correlations still remain.
We train a BERT-base classifier \cite{devlin2019bert} on the training set using only the claim text as input and find that it achieves a test accuracy of 75\%.
Since meaningful predictions without considering the prior art are impossible, this classifier must be relying on shortcuts.
To identify these shortcuts, we use an interpretable logistic regression model in \Cref{sec:results}.

To be able to determine whether a model relies on these shortcuts, we create an adversarial test set using adversarial filtering \cite{le2020adversarial} where models overfitted to these spurious correlations fail.
Specifically, we filter the test set, keeping only samples that are misclassified by the trained BERT model.
To ensure comparability with results from the full test set, we balance the dataset by randomly removing samples from the majority class.
Since it contains only 278 claims, it is not well-suited to evaluate and compare models in detail, but a substantial decrease in accuracy compared to the full test set is a strong indicator of overfitting to spurious correlations.

\subsection{Data Contamination Analysis}\label{sec:data_contamination}

Data contamination is a common concern when evaluating LLMs on domain-specific tasks, especially when the dataset is created from publicly available documents that may have been included in the pre-training data \cite{balloccu-etal-2024-leak,li-etal-2024-open-source,deng-etal-2024-investigating}.
We use \textsc{infini-gram} \cite{liu2024infinigram} to search for contents of our dataset in the largest indexed corpus,\footnote{\texttt{v4\_olmo-2-0325-32b-instruct\_llama}} which was used to train OLMo 2 \cite{walsh2025} and contains 4.6T tokens across 3M documents.
Our analysis shows that the relevant public data of our benchmark is not leaked into the largest openly indexed pre-training corpus.
First, we search for the phrase \enquote{\textit{The subject-matter of claim}} which appears verbatim in the majority of ESOP documents.
Only 58 documents in the index contain this phrase, the majority of which are EPO case law proceedings,\footnote{e.g. \url{https://www.epo.org/en/boards-of-appeal/decisions/t130520eu1}} and none of them are ESOP documents.
Second, we search for the exact text of the claims in our dataset.
We sample 100 claims, search for the first 5 words of each claim to avoid false negatives due to whitespace and punctuation differences, and manually verify whether matches contain the claim text.
Of the 100 sampled claims, none appear in the index; in many cases, even the first 5 words yield zero matches, despite consisting of generic phrases such as \textit{\enquote{An apparatus for communicating in.}}
Future work should conduct further analysis like time-based data splits \citep{knappich-etal-2025-pap2pat} to definitively rule out data contamination for models with closed pre-training corpora.

\begin{table*}
  \centering
  % \small
  \setlength{\tabcolsep}{.34em}
  \begin{tabular}{lcccccccccccccccccccc}
    \toprule

    \multirow{3}{*}{Model} &
    \multicolumn{14}{c}{Passage Retrieval (N=932)} & 
    \multicolumn{3}{c}{\multirow{2}{*}{\shortstack{Novel Feature\\Identification\\(N=912)}}} &
    \\
    
    \cmidrule(lr){2-15}

    &
    \multicolumn{7}{c}{Claim-level} &
    \multicolumn{7}{c}{Feature-level} &
    & & &
    \\

    \cmidrule(lr){2-8}
    \cmidrule(lr){9-15}
    \cmidrule(lr){16-18}

    &
    P &
    \~P &
    R &
    \~R &
    F1 &
    \~F1 &
    nDCG &
    P &
    \~P &
    R &
    \~R &
    F1 &
    \~F1 &
    nDCG &
    P &
    R &
    F1 &
    \\

    \midrule

    \multicolumn{19}{c}{\textit{Baselines}} \vspace{0.3em} \\

    Random &
    6.1 & 28.9 & 16.5 & 45.7 & 7.9 & 34.4 & 10.8 & 2.3 & 20.7 & 3.3 & 25.9 & 2.2 & 22.4 & 2.8 & 4.2 & 5.5 & 3.2 \\

    Rouge-L &
    11.3 & 43.1 & 25.5 & 55.2 & 13.6 & 46.9 & 17.7 & 4.6 & 26.6 & 11.9 & 34.3 & 5.7 & 28.9 & 7.9 & 32.2 & 36.8 & 30.9 \\

    Qwen3-8B-Embeddings &
    13.7 & 40.2 & 35.7 & 58.4 & 17.4 & 45.8 & 23.5 & 8.3 & 28.9 & 22.7 & 42.3 & 10.6 & 32.9 & 15.4 & 27.0 & 12.6 & 13.5 \\

    \midrule

    \multicolumn{19}{c}{\textit{Single-step Examination}} \vspace{0.3em} \\

    % Qwen3-4B-It &
    % 14.8 & 34.0 & 27.7 & 47.2 & 16.7 & 37.9 & 20.9 & 7.7 & 21.8 & 16.4 & 31.5 & 9.1 & 24.6 & 12.3 & 48.3 & 59.7 & 48.3 \\

    % \hspace{1em} w/ GRPO &
    % 16.1 & 38.9 & 42.3 & 63.0 & 21.1 & 46.5 & 28.0 & 9.3 & 26.9 & 24.5 & 42.7 & 11.9 & 31.6 & 17.0 & 44.4 & 37.9 & 36.9 \\

    % \hspace{1em} w/ GRPO + Think &
    % 21.3 & 45.1 & 38.4 & 60.3 & 24.1 & 49.8 & 30.5 & 13.6 & 31.5 & 25.0 & 42.5 & 15.3 & 34.7 & 19.6 & 37.2 & 32.6 & 30.6 \\

    Qwen3-30B-A3B-It &
    21.8 & 44.7 & 39.8 & 60.9 & 24.9 & 49.8 & 31.6 & 13.2 & 31.3 & 27.7 & 44.9 & 16.0 & 35.6 & 22.1 & 42.3 & 44.0 & 39.1 \\

    Qwen3-30B-A3B-Th &
    \textbf{22.7} & 45.3 & 41.6 & 60.6 & 25.4 & 49.8 & 31.8 & 12.2 & 26.6 & 26.5 & 38.8 & 14.7 & 30.3 & 19.9 & 42.5 & 44.9 & 38.6 \\

    Qwen3-VL-235B-A22B-Th &
    20.4 & 44.0 & 52.4 & 70.3 & 26.1 & 52.3 & 35.1 & 13.8 & 32.4 & 36.0 & 52.1 & 17.7 & 38.4 & 25.1 & 43.4 & 47.2 & 41.3 \\

    Qwen3.5-397B-A17B &
    20.2 & 44.9 & \textbf{57.6} & \textbf{74.8} & 27.1 & 54.4 & \textbf{36.6} & 14.8 & 34.7 & 38.9 & 55.6 & 19.2 & 41.1 & 27.3 & 45.2 & 52.7 & 44.7 \\

    % \hspace{1em} w/ self-consistency &
    % 22.3 & \textbf{46.6} & 53.6 & 71.6 & \textbf{28.2} & 54.7 & 35.9 & 16.9 & 36.4 & 36.5 & 52.9 & 20.4 & 41.3 & 27.2 & 46.1 & 54.4 & 46.0 \\

    \midrule

    \multicolumn{19}{c}{\textit{Hierarchical Examination}} \vspace{0.3em} \\

    Qwen3-30B-A3B-Th &
    21.5 & \textbf{46.3} & 44.9 & 65.1 & 25.5 & 52.2 & 33.6 & 12.8 & 29.6 & 26.2 & 41.6 & 14.9 & 33.2 & 19.7 & 48.5 & 73.8 & 53.7 \\

    \hspace{1em} w/ LLM segmentation &
    21.4 & \textbf{46.3} & 44.4 & 64.7 & 25.2 & 52.1 & 32.7 & 12.7 & 28.8 & 27.1 & 41.2 & 15.0 & 32.4 & 19.9 & 50.9 & 56.1 & 48.2 \\

    Qwen3.5-397B-A17B &
    20.7 & 46.2 & 56.8 & 74.2 & \textbf{27.4} & \textbf{55.1} & 35.5 & \textbf{17.0} & \textbf{38.4} & \textbf{39.1} & \textbf{57.3} & \textbf{20.9} & \textbf{44.1} & \textbf{28.6} & \textbf{53.5} & \textbf{81.8} & \textbf{59.7} \\

    \bottomrule
\end{tabular}
  \caption{
    Experimental Results for \textit{Passage Retrieval} and \textit{Novel Feature Identification}. 
    \textit{Passage Retrieval} metrics are only computed for rejected samples and \textit{Novel Feature Identification} metrics are only computed for granted claims. 
    \~P, \~R, and \~F1 denote the soft variants of precision, recall, and F1. 
    Best results are in \textbf{bold}.
  }
  \label{tab:results_retrieval}
\end{table*}

\section{LLM Workflows}\label{sec:methods}

In this section, we describe the LLM workflows that we implement to perform automatic patent novelty prediction with passage retrieval at the claim and feature level.
An overview of the workflows is shown in \Cref{fig:workflows}.
Each workflow is implemented using DSPy \cite{khattab2024dspy} and generates a structured output containing (i) the relevant prior art passages for each feature of the claim, (ii) a novelty prediction for each feature (\textit{\enquote{Fully Disclosed}, \enquote{Partially Disclosed}, \enquote{Not Disclosed}}), and (iii) an overall novelty prediction for the claim (\textit{\enquote{Novel}}, \textit{\enquote{Not Novel}}).

\paragraph{Single Step Examination}

In the \textit{Single Step Examination} workflow, the model receives the claim and prior art document as input and is prompted to output the structured response in a single step.
On average, the prompt contains 17k tokens.

\paragraph{Hierarchical Examination}

In the \textit{Hierarchical Examination} workflow, each feature is analyzed separately, allowing the model to perform deeper reasoning on each feature. It follows three main steps:

\begin{enumerate}[wide,labelindent=0pt]
  \item Claim Segmentation: First, the claim is split into individual features either using a dedicated LLM call or by splitting at semicolons and newlines.
  \item Feature-level Retrieval \& Examination: Next, for each feature, the LLM is called with the feature, the full claim for context, and the prior art document. The model analyzes the prior art document with respect to the given feature and outputs the identifiers of relevant prior art passages, a novelty prediction, and an examination summary for the given feature.
  \item Claim-level Novelty Prediction: Finally, the model is called with the claim under examination, the prior art document filtered according to the retrieval results of the previous step, and the feature-level examination summaries. The model outputs the overall novelty prediction for the claim.
\end{enumerate}

This approach requires a total of 129k prompt tokens per claim on average (over 7x of \textit{Single Step Examination}), with 95\% of the tokens used for feature-level examination.
However, each feature's examination can be performed in parallel.
We order the elements in the prompt for feature-level examination in a way that maximizes prefix cache hit rate (i.e., prior art, claim, feature).
Overall, despite requiring over 7x more tokens, it runs only 2.5x longer than \textit{Single Step Examination} on a dedicated vLLM deployment with prefix caching.

\section{Experiments}\label{sec:experiments}

In this section, we describe our experimental setup, including several baselines (\Cref{sec:baselines}) and results (\Cref{sec:results}).
We focus our analysis on open-weight models because patent examination requires high standards of data privacy and security that are difficult to guarantee when using closed-weight models via API.
We deploy \texttt{Qwen/Qwen3\--30B\--A3B\--Thinking\--2507} \cite{qwen3}, \texttt{Qwen/Qwen3\--VL\--235B\--A22B\--Thinking\--FP8} \cite{Qwen3-VL}, and \texttt{Qwen/Qwen3.5\--397B\--A17B\--FP8}\footnote{\url{https://huggingface.co/Qwen/Qwen3.5-397B-A17B-FP8}} via vLLM \cite{kwon2023efficient}.

\begin{figure*}[t]
  \centering
  \input{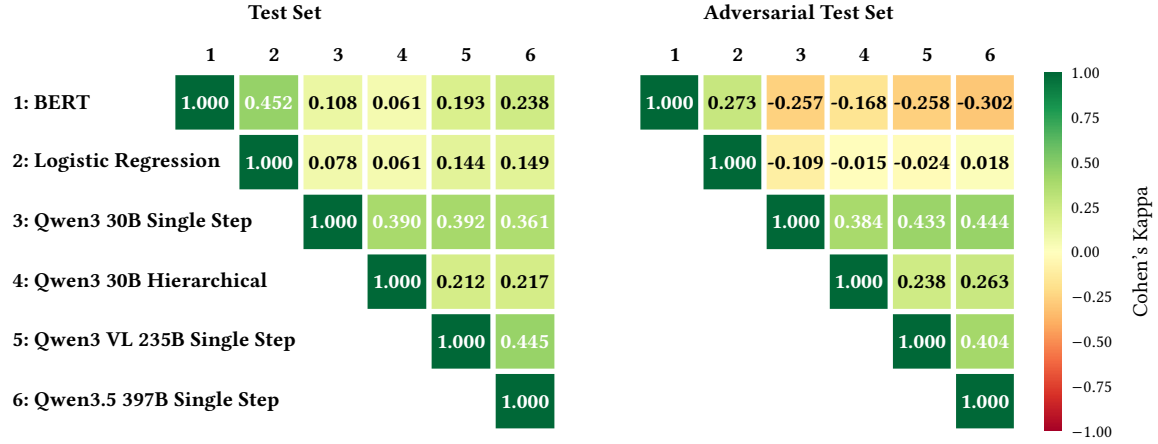}
  \caption{
    Cohen's kappa agreement between different models' predictions on the test and adversarial test sets. 
  }
  \Description{Cohen's kappa agreement between different models' predictions on the test and adversarial test sets.}
  \label{fig:cohen_kappa}
\end{figure*}

\subsection{Baselines}\label{sec:baselines}

We implement several baselines for a better understanding of the task and metrics: \textit{Random} across all sub-tasks, \textit{Logistic Regression} and \textit{BERT Classifier} for claim-level novelty classification, and \textit{Embedding Similarity} and \textit{ROUGE-L} for passage retrieval and novel feature identification.

\subsubsection{Random}

To establish a lower bound, we implement a random baseline that assigns claim-level and feature-level novelty labels and retrieves passages uniformly at random.

\subsubsection{ROUGE-L}

To establish a simple baseline for passage retrieval and novel feature identification using only lexical overlap, we compute the ROUGE-L score \cite{lin-2004-rouge} between each feature and each passage in the prior art document.
We consider a feature to be disclosed if any passage has a similarity score above 0.4.

\subsubsection{Embedding Similarity}

In addition, we implement an embedding-based baseline for passage retrieval and novel feature identification using \texttt{Qwen/Qwen3\--Embedding\--8B} \cite{qwen3embedding}.
Similar to the ROUGE-L baseline, we compute the cosine similarity between each feature and each passage in the prior art document and consider a feature to be disclosed if any passage has a similarity score above 0.5.

\subsubsection{Logistic Regression}

To analyze the spurious correlations in the dataset, we implement a logistic regression model using various features based solely on the claim under examination without considering the prior art document. 
These features include the number of words, the number of features, the assigned IPC\footnote{\url{https://www.wipo.int/en/web/classification-ipc}} classes (indicating the technical domain), and the top 500 TF-IDF features with n-grams up to length 4. 
We use a standard feature scaler to ensure that the model coefficients are interpretable.
This baseline is only applied to the claim-level novelty classification task.

\subsubsection{BERT Classifier}

Like the logistic regression model, this baseline only considers the claim under examination without the prior art document, and should thus not be able to make any meaningful predictions.
We fine-tune a BERT-base model \cite{devlin2019bert} on the training set for claim-level novelty classification for 3 epochs.

\subsection{Results}\label{sec:results}

\Cref{tab:results_retrieval} shows the results for \textit{Passage Retrieval} and \textit{Novel Feature Identification}, and \Cref{tab:results_cls} shows the results for \textit{Claim Novelty Prediction}.

\subsubsection{Passage Retrieval}

The LLM workflows clearly outperform lexical and semantic similarity baselines across all passage retrieval metrics. 
At both claim-level and feature-level, the best-performing model is Qwen3.5-397B-A17B using hierarchical examination, achieving a claim-level F1 score of 27.4 and a feature-level F1 score of 20.9 compared to 17.4 and 10.6 respectively for Qwen3-8B-Embeddings. 
This represents a substantial relative improvement of over 50\% over the strongest non-LLM baseline.
Hierarchical examination yields moderate but consistent retrieval improvements for Qwen3.5-397B-A17B, whereas Qwen3-30B-A3B-Thinking shows no clear benefit, suggesting that the advantages of task decomposition for retrieval may grow with a model's capabilities.
Overall, performance differences across configurations are more pronounced at the feature level than at the claim level, confirming that fine-grained evaluation provides a more discriminative signal for comparing models.
% Sampling reasoning multiple reasoning trajectories and averaging the results (self-consistency) yields improvements in claim-level and feature-level F1, but not in nDCG.
Finally, retrieval performance scales consistently with model size across all three models.

\subsubsection{Novel Feature Identification}

All LLM workflows substantially outperform the baselines on novel feature identification, even in the single-step setting.
Notably, hierarchical examination yields large gains over single step examination, with Qwen3-30B-A3B-Thinking improving from 38.6 to 53.7 F1 and Qwen3.5-397B-A17B from 44.7 to 59.7 F1.
Furthermore, hierarchical examination with the 30B model outperforms single-step examination with the larger 397B model (53.7 vs.\ 44.7 F1), revealing that structured decomposition is more critical than model scale for this sub-task.

\subsubsection{Claim Novelty Prediction}

\paragraph{Fraction of Predicted Novel Claims}

A notable observation across models is the significant variation in the fraction of claims predicted as novel. 
The Random baseline predicts approximately 49.5\% of claims as novel, closely matching the balanced dataset distribution. 
In contrast, Qwen3-30B-A3B-Instruct (without thinking mode) exhibits extreme bias and predicts merely 3.2\% as novel, indicating a strong tendency to classify claims as lacking novelty when not engaging in extended reasoning.
The reasoning-enabled models demonstrate more balanced prediction distributions.
Using single-step examination, Qwen3-30B-A3B-Thinking, Qwen3-VL-235B-A22B-Thinking, and Qwen3.5-397B-A17B predict 69.1\%, 48.8\%, and 50.8\% as novel.
Using hierarchical examination, the predicted novel fractions vary widely across models (30\%--88\%), suggesting that the effect of task decomposition on prediction bias is model-dependent.
Overall, these variations show that even minor changes to the setup can substantially impact model behavior, underscoring the importance of comprehensive evaluation and calibration in real-world applications.

\paragraph{Spurious Correlations}

On the test set, the fine-tuned BERT classifier achieves the highest accuracy of 75.2\%, despite not having access to the prior art document.
This indicates that, even after stratifying the dataset based on claim length and removing reference numerals, subtle but effective spurious correlations remain that models can exploit to achieve high claim novelty prediction performance without performing the intended task.
The logistic regression model captures a substantial portion of BERT's performance gain over random guessing, suggesting that BERT relies heavily on simple, interpretable features.
Although this does not definitively prove that BERT uses identical features to the logistic regression model, their predictions show moderate agreement (Cohen's Kappa of 0.45).
We inspect the learned weights of the logistic regression model, and find that among the most important features are (i) the presence of specific n-grams that are used to narrow a claim's scope (e.g., \enquote{\textit{wherein}}, \enquote{\textit{characterized}}, \enquote{\textit{in that the}}), (ii) the indicator variables for certain IPC classes (technical domains) with imbalanced label distributions, (iii) the number of features in the claim, and (iv) the counts of punctuation characters.

\paragraph{LLMs' Robustness to Spurious Correlations}

Interestingly, we find that the implemented LLM workflows do not appear to exploit these spurious correlations.
As described in \Cref{sec:dataset_spurious}, we create an adversarial test set using adversarial filtering to identify samples that are misclassified by the BERT classifier.
On this adversarial subset, the LLM workflows show only minor accuracy loss or even gains compared to the full test set.
We note that due to the small size of the adversarial test set (278 samples), these results have very high variance.
Therefore, we propose not to use this subset to evaluate the performance of models directly, but rather to verify that models do not exploit the identified spurious correlations.
Additionally, we compute the agreement between different models' predictions using Cohen's Kappa \cite{cohen1960coefficient}, as shown in \Cref{fig:cohen_kappa}.
LLM workflows show only low to moderate agreement with the BERT classifier and the logistic regression model on the test set, and negative agreement on the adversarial test set, further indicating that they do not rely on the same spurious correlations.

\paragraph{Cross-Dataset Analysis}

To validate that the issue of spurious correlations is not specific to our dataset, we conduct a similar analysis on the NOC4PC task of the PANORAMA dataset \cite{lim2025panorama}. 
We train an XGBoost model on their claim novelty prediction task (NOC4PC) using TF-IDF and structural features from the claim text alone.
The model achieves a test macro F1 of 41\%, substantially outperforming most prompted LLMs in their evaluation.  
This confirms that spurious correlations are a prevalent issue in patent novelty datasets and that future work should carefully validate that models do not exploit them.

\begin{table}[t]
  \centering
  % \small
  \setlength{\tabcolsep}{.4em}
  \begin{tabular}{lccc}
    \toprule

    Model &
    \shortstack{Predicted\\Novel} &
    Accuracy &
    Macro F1
    \\

    \midrule

    \multicolumn{4}{c}{\textit{Baselines}} \vspace{0.3em} \\

    Random &
    49.5\% & 50.1 / 49.6 & 50.1 / 49.6 \\

    Logistic Regression &
    47.7\% & 66.3 / 36.3 & 66.2 / 36.3 \\

    BERT (claim only) &
    39.7\% & \textbf{75.2} / 0.0 & \textbf{74.9} / 0.0 \\
    \midrule

    \multicolumn{4}{c}{\textit{Single-step Examination}} \vspace{0.3em} \\

    % Qwen3-4B-It &
    % 0.2\% & 50.6 / 50.4 & 33.6 / 33.5 \\

    % \hspace{1em} w/ GRPO &
    % 0.1\% & 50.5 / 49.6 & 33.7 / 33.2 \\

    % \hspace{1em} w/ GRPO + Think &
    % 35.4\% & 61.5 / 60.7 & 60.6 / 59.2 \\

    Qwen3-30B-A3B-It &
    3.2\% & 52.0 / 51.1 & 38.2 / 36.9 \\

    Qwen3-30B-A3B-Th &
    69.1\% & 61.4 / 62.2 & 60.0 / 60.9 \\

    \hspace{1em} w/o prior art &
    75.2\% & 53.7 / 55.0 & 50.7 / 52.3 \\

    Qwen3-VL-235B-A22B-Th &
    48.8\% & 65.1 / 62.8 & 65.1 / 62.8 \\

    Qwen3.5-397B-A17B &
    50.8\% & 69.1 / 65.1 & 69.1 / 65.1 \\

    \hspace{1em} w/ self-consistency &
    53.2\% & 70.7 / \textbf{67.5} & 70.7 / \textbf{67.5} \\

    \midrule

    \multicolumn{4}{c}{\textit{Hierarchical Examination}} \vspace{0.3em} \\

    Qwen3-30B-A3B-Th &
    88.5\% & 57.9 / 57.2 & 50.8 / 49.8 \\

    \hspace{1em} w/o summaries &
    75.3\% & 61.5 / 61.7 & 59.1 / 59.0 \\

    \hspace{1em} w/ label references &
    69.0\% & 61.6 / 58.3 & 60.3 / 56.9 \\

    Qwen3.5-397B-A17B &
    30.2\% & 63.3 / 65.5 & 61.8 / 64.0 \\

    \bottomrule
\end{tabular}
  \caption{Experimental Results for \textit{Claim Novelty Prediction}. For accuracy and macro F1, we report the performance on the full test set and on the adversarial subset (after the slash). Best results are in \textbf{bold}.}
  \label{tab:results_cls}
\end{table}

\paragraph{Prediction Accuracy}

While Qwen3-30B-A3B-Instruct performs close to random chance, the thinking-enabled models achieve substantially better results. 
Using single-step examination, Qwen3-30B-A3B-Thinking, Qwen3-VL-235B-A22B-Thinking, and Qwen3.5-397B-A17B achieve 61.4\%, 65.1\%, and 69.1\% accuracy respectively, representing meaningful improvements over the random baseline of 50.1\%.
Sampling multiple reasoning trajectories and averaging the results (self-consistency) yields further improvements in accuracy, with the 397B model achieving 70.7\% accuracy.
Hierarchical examination does not aid claim-level novelty prediction, i.e., the added reasoning effort does not translate into better claim-level predictions.
Furthermore, adding feature examination summaries actually reduces accuracy to 57.9\% and makes predictions even more unbalanced (88.5\% predicted as novel), suggesting that the intermediate summaries may introduce noise or bias into the final aggregation step.
Even using the label references from the ESOP document directly (i.e., using an oracle for the retrieval step; see \Cref{tab:results_cls} \enquote{with label references}) does not improve claim novelty prediction accuracy, indicating that passage retrieval is not the main bottleneck for the binary prediction.
Overall, while absolute accuracies leave substantial room for improvement, the robustness of LLMs against spurious correlations is particularly encouraging for practical deployment.

\section{Discussion and Limitations}

Our experimental results demonstrate both the promise and challenges of automated patent novelty examination at the feature level.
While LLM-based workflows substantially outperform embedding-based baselines on passage retrieval and novel feature identification, absolute performance levels indicate room for improvement.
Our multi-faceted evaluation reveals that different design choices yield varied improvements across evaluation facets: hierarchical examination enhances passage retrieval and novel feature identification but hurts claim-level novelty prediction, whereas model scale consistently improves performance across all facets.

\paragraph{Reliability of ESOP Documents}

ESOP documents reflect the immense effort of highly trained patent examiners.
In particular, EPO examination records are widely considered to be among the highest quality, given that each application is, unlike in other jurisdictions, examined by at least three examiners in order to reach a joint decision.\footnote{\url{https://www.epo.org/en/legal/epc/2020/a18.html}}
Nevertheless, they are not infallible and may contain errors or omissions, as exemplified by opposition procedures.
Given the complexity and technical depth of patent examination, future work should further investigate how reliable ESOP documents are as a source of supervision, how a team of expert humans would perform on the proposed tasks, and how large their inter-annotator agreement would be.

\paragraph{Extension to Multi-Document Settings}

The examination of novelty is formally always performed with respect to a single prior art document, thus our work has focused on this setting.
However, it is likely that examiners perform such analysis across multiple documents, and record only the best match.
Future work should explore extending the proposed approaches to multi-document settings, where workflows must first identify relevant documents, perform novelty examination with respect to each document, and aggregate results into a final examination report.

\paragraph{Generalizability Across Domains}

While our work focuses on patents from specific technical domains (computing, speech processing, telecommunications, and pictorial communication) filed at the EPO, the proposed methodology is applicable more broadly.
The feature-level examination approach is domain-agnostic and could be extended to other technical fields, though domain-specific language understanding may present additional challenges.
Furthermore, while we focus exclusively on patent literature as prior art, novelty examination in practice also considers non-patent literature such as scientific publications, technical manuals, and public disclosures.
Incorporating such diverse sources would require addressing challenges in document retrieval, format heterogeneity, and citation extraction, but would enable more comprehensive novelty assessment systems.

\section{Conclusion}

In this work, we have challenged the prevailing paradigm of treating patent novelty examination as a simple binary classification task. 
We introduced \DatasetName, a novel dataset comprising 3,658 first patent claims with fine-grained, feature-level prior art references extracted from European Search Opinion documents. 
This dataset enables evaluation at fine granularity to assess models' ability to identify specific passages disclosing individual claim features and to recognize which features make a claim novel.
Our analysis revealed significant spurious correlations in existing approaches to novelty prediction, including claim length, reference numerals, and domain-specific linguistic patterns. 
We demonstrated that trained classifiers readily exploit these shortcuts, achieving high accuracy without genuinely reasoning about claim-prior art relationships. 
To address this, we created an adversarial test set using adversarial filtering, providing a diagnostic tool to verify model robustness.
We proposed and evaluated LLM workflows that decompose claims into features, analyze each feature against prior art, and derive claim-level predictions. 
Our experiments demonstrated that these workflows substantially outperform embedding-based baselines on passage retrieval and novel feature identification tasks. 
Crucially, unlike fine-tuned classifiers, LLMs proved robust against spurious correlations, maintaining consistent performance on the adversarial test set.
Our findings have important implications for both the research community and patent examination practice. 
For researchers, we provide a challenging benchmark that requires genuine semantic understanding and abstract reasoning, along with fine-grained supervision enabling more nuanced model development. 
For practitioners, our work points toward AI systems that can provide transparent, verifiable assistance to patent examiners, patent attorneys, and domain experts by explicitly linking predictions to supporting evidence in prior art documents.
We release the \DatasetName dataset, the dataset creation pipeline, and all experimental code to foster further research into transparent and granular patent analysis. 
Future work should explore extending these approaches to multi-document retrieval settings, designing robust training pipelines, and establishing human expert performance.

% \begin{acks}
% \end{acks}

%%
%% The next two lines define the bibliography style to be used, and
%% the bibliography file.
\bibliographystyle{ACM-Reference-Format}
\bibliography{references.bib}

%%
%% If your work has an appendix, this is the place to put it.
% \appendix

% \section{Test}\label{sec:appendix_test}

\end{document}